\documentclass[lettersize,journal]{IEEEtran}
\usepackage{amsmath,amsfonts}
\usepackage{algorithmic}
\usepackage{algorithm}

\usepackage{array}
\usepackage[caption=false,font=normalsize,labelfont=sf,textfont=sf]{subfig}
\usepackage{textcomp}
\usepackage{stfloats}
\usepackage{url}
\usepackage{verbatim}
\usepackage{graphicx}
\usepackage{cite}
\usepackage{hyperref}
\usepackage{booktabs}
\usepackage{bm}
\usepackage{multirow}
\usepackage{tabularx}
\usepackage{orcidlink}
\begin{document}

\title{An Iteration-Free Fixed-Point Estimator for Diffusion Inversion}

\author{
Yifei Chen\orcidlink{0009-0002-8725-8401}, 
Kaiyu Song\orcidlink{0009-0007-6443-7104},
Yan Pan\orcidlink{0000-0002-0466-3763},
Jianxing Yu\orcidlink{0000-0003-1340-3995}~\IEEEmembership{Member,~IEEE},
Jian Yin\orcidlink{0000-0002-1214-5384}~\IEEEmembership{Member,~IEEE},
and
Hanjiang Lai\orcidlink{0000-0001-8057-6744}~\IEEEmembership{Member,~IEEE}

\thanks{Hanjiang Lai is the corresponding author.}
\thanks{Yifei Chen, Yan Pan and Hanjiang Lai are with the School of Computer Science and Engineering, Sun Yat-sen University, Guangzhou 510006, China (e-mail: chenyf387@mail2.sysu.edu.cn, panyan5@mail.sysu.edu.cn, laihanj3@mail.sysu.edu.cn).}
\thanks{Kaiyu Song, Jianxing Yu and Jian Yin are with the School of Artificial Intelligence, Sun Yat-sen University, Zhuhai 519082, China (e-mail: songky7@mail2.sysu.edu.cn, yujx26@mail.sysu.edu.cn, issjyin@mail.sysu.edu.cn).}
}

\markboth{IEEE TRANSACTIONS ON MULTIMEDIA}%
{Shell \MakeLowercase{\textit{et al.}}: A Sample Article Using IEEEtran.cls for IEEE Journals}

\IEEEpubid{0000--0000/00\$00.00~\copyright~2021 IEEE}

\maketitle

\begin{abstract}
Diffusion inversion aims to recover the initial noise corresponding to a given image such that this noise can reconstruct the original image through the denoising diffusion process. The key component of diffusion inversion is to minimize errors at each inversion step, thereby mitigating cumulative inaccuracies. Recently, fixed-point iteration has emerged as a widely adopted approach to minimize reconstruction errors at each inversion step. However, it suffers from high computational costs due to its iterative nature and the complexity of hyperparameter selection. To address these issues, we propose an iteration-free fixed-point estimator for diffusion inversion. First, we derive an explicit expression of the fixed point from an ideal inversion step. Unfortunately, it inherently contains an unknown data prediction error. Building upon this, we introduce the error approximation, which uses the calculable error from the previous inversion step to approximate the unknown error at the current inversion step. This yields a calculable, approximate expression for the fixed point, which is an unbiased estimator characterized by low variance, as shown by our theoretical analysis. We evaluate reconstruction performance on two text-image datasets, NOCAPS and MS-COCO. 
Compared to DDIM inversion and other inversion methods based on the fixed-point iteration, our method achieves consistent and superior performance in reconstruction tasks without additional iterations or training. 
\end{abstract}

\begin{IEEEkeywords}
Diffusion Models, Diffusion Inversion, Fixed-Point Iteration.
\end{IEEEkeywords}

\section{Introduction}


\IEEEPARstart{D}{iffusion} inversion~\cite{AIDI,garibi2024renoise,mokady2023null,dong2023prompt,hong2024exact,wallace2023edict,wang2024belm,zhang2024exact,zhang2024easyinv} aims to recover the initial latent noise corresponding to a given image. This noise, which is taken as the starting point for the model's denoising trajectory, should enable a high-fidelity reconstruction of the original image. As a pivotal technique, diffusion inversion greatly extends the utility of diffusion models.  It enables a wide range of downstream tasks, 
such as prompt-to-prompt image editing~\cite{tumanyan2023plug, hertz2022prompt}, style transfer~\cite{chung2024style}, high-resolution image generation~\cite{vontobel2025hiwave}, and direct preference optimization (DPO)~\cite{li2025inversion}.

\begin{figure*}
	\begin{minipage}[b]{0.33\linewidth}
		\centering
		\subfloat[\label{fig:ideal_inv_diagram}]{\includegraphics[width=0.9\linewidth]{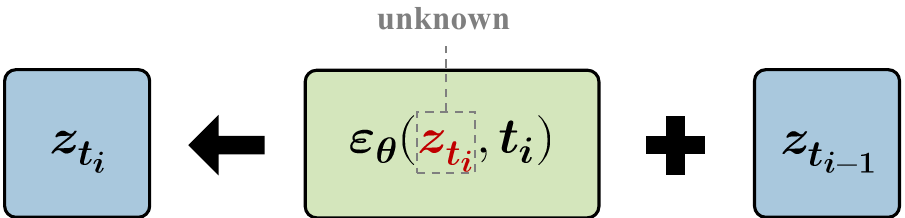}}

		\subfloat[\label{fig:ddim_inv_diagram}]{\includegraphics[width=0.9\linewidth]{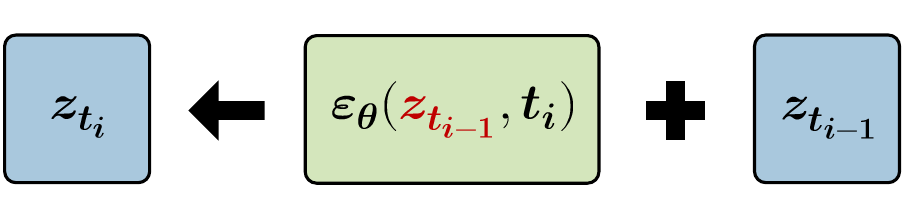}}
    
	\end{minipage}
	\begin{minipage}[b]{0.33\linewidth}
		\centering
		\subfloat[\label{fig:fixed_point_diagram}]{\includegraphics[width=0.9\linewidth]{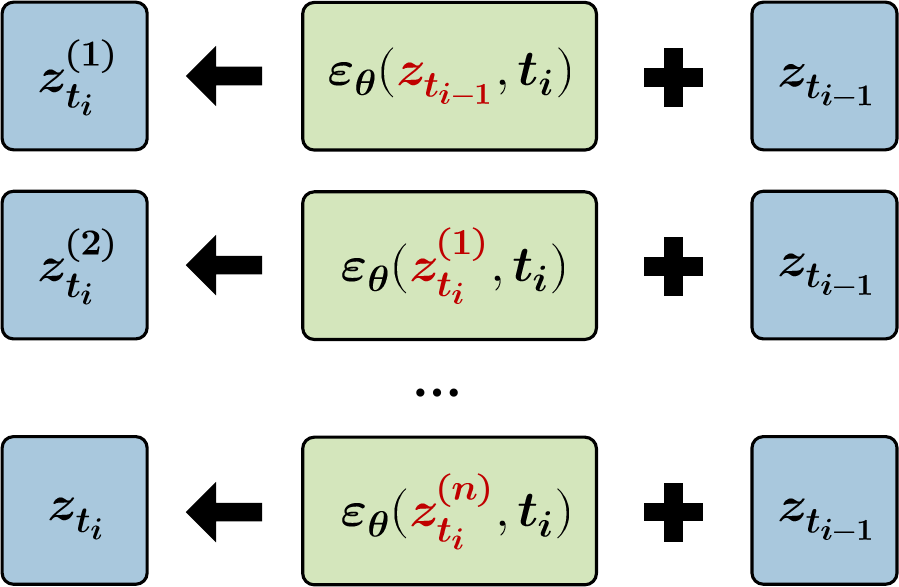}}
    
	\end{minipage}
 \begin{minipage}[b]{0.33\linewidth}
		\centering
		\subfloat[\label{fig:our_diagram}]{\includegraphics[width=0.9\linewidth]{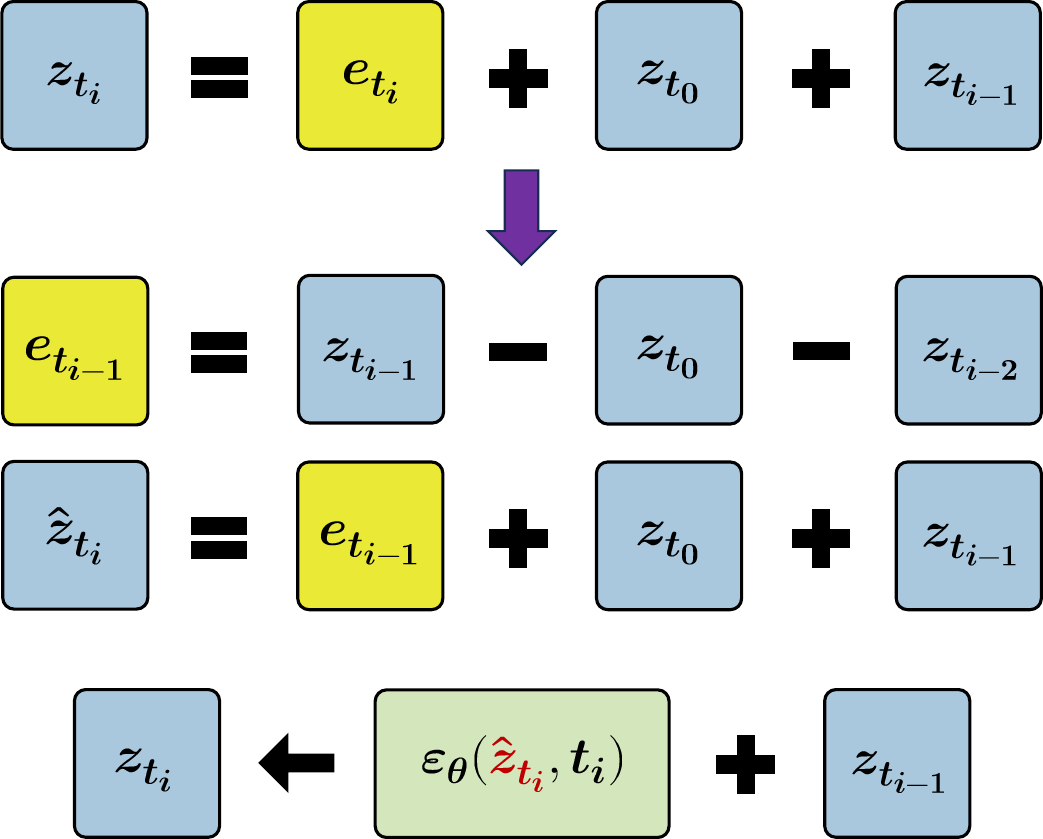}}
    
	\end{minipage}
	\caption{(a) One ideal inversion step, whose input latent of the neural network is unknown. (b) One inversion step of DDIM Inversion, which approximates the input latent $\bm{z}_{t_i}$ by the preceding latent $\bm{z}_{t_{i-1}}$. (c) One inversion step of fixed-point-iteration-based methods, which includes additional iterations. (d) One inversion step of our method, which estimates the fixed point without iterations and inputs the estimate to the neural network.}
\end{figure*}

Diffusion inversion~\cite{AIDI,garibi2024renoise} is performed by leveraging the pre-trained diffusion models for denoising and generating.
While the standard denoising process advances $\bm{z}_{t_i}$ to $\bm{z}_{t_{i-1}}$, diffusion inversion proceeds in the opposite direction from $\bm{z}_{t_{i-1}}$ to $\bm{z}_{t_i}$. This introduces a key challenge: at each inversion step, the method requires access to the latent variable $z_t$, which is unknown during inversion, as illustrated in Fig.~\ref{fig:ideal_inv_diagram}.
DDIM inversion~\cite{ddim} relies on the assumption that the pre-trained diffusion model yields approximately consistent score predictions across adjacent timesteps. 
It approximates the unknown latent $\bm{z}_{t_i}$ by the preceding latent $\bm{z}_{t_{i-1}}$ (see Fig. \ref{fig:ddim_inv_diagram}). Such an approximation introduces an error at each inversion step, which leads to the accumulation of error throughout the entire inversion-denoising process.

To mitigate accumulated errors, several methods~\cite{jupnp,mokady2023null,dong2023prompt} redesign the denoising trajectory to achieve higher reconstruction accuracy. This strategy 
adapts the denoising process itself to ensure its alignment with the DDIM inversion process. 
For example, Ju et al.~\cite{jupnp} use the discrepancy between the inversion and denoising latents as a corrective signal, which they integrate into the denoising process to align it with the inversion trajectory.
Different from the above methods, another line of inversion methods~\cite{zhang2024easyinv, AIDI, garibi2024renoise} achieves accurate reconstruction without modifying the denoising process.
They minimize the error introduced at each inversion step to mitigate the issue of error accumulation and recover the denoising process.
The representative method is fixed-point-iteration-based methods~\cite{AIDI,garibi2024renoise} (see Fig. \ref{fig:fixed_point_diagram}). 
For instance, AIDI \cite{AIDI} adopts the fixed-point iteration process for each inversion step and accelerates the iterative process. 
ReNoise \cite{garibi2024renoise} also applies fixed-point iteration to diffusion inversion, which accelerates convergence and yields more editable initial noise by refining both the latent and the predicted noise. 

\IEEEpubidadjcol
These fixed-point iteration methods mitigate per-step error accumulation in the DDIM inversion process, whereas it still faces two well-known limitations. First, it requires multiple iterative updates in each step, leading to prohibitively high computational overhead during inversion. Second, as empirically observed, the reconstruction quality converges in an oscillatory manner with increasing iterations per step. Consequently, when the iteration count is selected within an insufficient range, the resulting performance gain may remain marginal or even be negative, relative to the additional computational cost incurred. 

To address these limitations, we introduce a novel and efficient diffusion inversion method, named Iteration-Free Fixed-Point Estimator (IFE). As shown in the top row of Fig. \ref{fig:our_diagram}, we first derive another explicit expression for the unknown latent $\bm{z}_{t_i}$, which is independent of the neural network.
However, this expression contains an unknown error in data prediction. Fortunately, the error term at the preceding inversion step can be computed (see the second row in Fig. \ref{fig:our_diagram}). Substituting the error term at the current inversion step with the one at the preceding inversion step yields a fixed-point estimator (see the third row in Fig. \ref{fig:our_diagram}). Furthermore, through both theoretical analysis and empirical validation, we also show that our fixed-point estimator is an unbiased, low-variance estimator with high computational efficiency. In summary, our proposed iteration-free fixed-point estimator achieves superior performance and higher efficiency than the fixed-point-iteration-based methods. 

The contributions can be summarized as follows:

\begin{itemize}
  \item We propose an explicit closed-form expression for the unknown latent $\bm{z}_{t_i}$ in diffusion inversion, and approximate the unknown prediction error in this expression by the error at the preceding inversion step. This formulation yields a high-quality estimate of the unknown latent without requiring any iterative refinement.
  \item We theoretically prove that this simple yet effective approximation constitutes an unbiased fixed-point estimator with low variance, establishing strong statistical guarantees for our method.
  \item Extensive experiments on two benchmark datasets show that our method achieves reconstruction performance competitive with state-of-the-art baselines.
\end{itemize}

\section{Related Work}
\subsection{Diffusion Models}
The advent of diffusion models has established a new and powerful paradigm for image generation and various tasks~\cite{tmm11207541, tmm11153996, tmm11071375, TMM10814072}. By learning to iteratively denoise data from a random distribution into a coherent sample, these models have achieved unprecedented levels of synthesis quality and diversity~\cite{ddpm, iddpm, ldm}. Song et al.~\cite{song2021maximum, song2020score} provided an insight that diffusion models can be elegantly generalized to a continuous-time framework using differential equations with the score function. Karras et al.~\cite{karras2022elucidating} established a unified framework integrating DDPM~\cite{ddpm} with score matching~\cite{song2020score}, which introduced an improved parameterization for the noise scheduling mechanism. This viewpoint provides a unified and mathematically rigorous foundation for understanding and improving sampling of diffusion models~\cite{ddim, lu2022dpm, zhao2023unipc}. By incorporating conditional terms into the score function, diffusion models support controlled synthesis through conditional guidance. Dhariwal et al.~\cite{ClassifierG} introduced classifier-guided diffusion models for conditional generation, whereas Ho et al.~\cite{CFG} proposed integrating conditions directly into the model during training to achieve classifier-free guided generation. 

\subsection{Inversion in Diffusion Models}
Generally, identifying the initial noise corresponding to a given image is crucial in diffusion inversion, since initiating the denoising process with an accurate noise estimate enables faithful reconstruction of all intermediate latents along the original denoising trajectory.
Some investigations have utilized fixed-point iteration methods to reduce single-step errors during inversion steps. AIDI~\cite{AIDI} first introduced fixed-point iteration for inversion process optimization, employing Anderson acceleration~\cite{anderson1965iterative} to improve both the speed and convergence stability of the iterative procedure. ReNoise~\cite{garibi2024renoise} extended the fixed-point iteration framework by incorporating gradient-based optimization of both the predicted and stochastic noise terms, thereby enabling compatibility with SDE sampling processes. Beyond methods based on fixed-point iteration, Lin et al.~\cite{lin2024schedule} introduced Schedule Your Edit, which employs a noise schedule specifically adapted for inversion to mitigate the singularity problem inherent in the inversion process. Chung et al.\cite{chung2024cfgpp} introduce a novel classifier-free guidance approach for image generation and demonstrate its efficacy in mitigating the excessive errors caused by large guidance scales in diffusion inversion.

An alternative strategy employed by a class of inversion techniques does not dedicate much effort to the search for an exact initial noise, but achieves an accurate reconstruction trajectory through the infusion of information from the inversion process into the denoising process. They leverage DDIM inversion for the inversion process and integrate information from the inversion process directly into the denoising process. Null-Text Inversion~\cite{mokady2023null} and Prompt-Tuning Inversion~\cite{dong2023prompt} align the latents of the denoising process with those of the inversion process by optimizing null-text embeddings and prompt embeddings, respectively. 
PnPInversion~\cite{jupnp} decouples the denoising process into target and source branches, and then aligns the latents of the source branch with those from the inversion process. It improves reconstruction fidelity while maintaining considerable editability. A common requirement across these approaches is the modification of the denoising procedure. This requirement restricts their scope of application, as it introduces compatibility issues and compromises their orthogonality with other denoising-time optimization algorithms. 

Furthermore, a range of inversion techniques building upon other solvers have been developed. Notable examples include EDICT~\cite{wallace2023edict}, BDIA~\cite{zhang2024exact}, and BELM~\cite{wang2024belm}, which employ a linear multistep method for highly precise reconstructions, and the DPM-Solver-based inversion method introduced by Hong et al.~\cite{hong2024exact}.

\subsection{Applications of Diffusion Inversion}
Diffusion inversion provides a foundational anchor for a wide range of creative and practical applications. A primary application is prompt-to-prompt image editing\cite{hertz2022prompt, cao2023masactrl, parmar2023zero, tumanyan2023plug}. By ensuring it is able to recover the denoising path with the inverted noise, these methods enable edits that are both faithful to the editing prompt and coherent with the original image's features.
Additionally, Chung et al. \cite{chung2024style} leverages diffusion inversion to extract features from both style and content images, thereby facilitating effective style transfer. Diffusion inversion is also adopted for other tasks. Vontobel et al.\cite{vontobel2025hiwave} propose the use of patch-wise DDIM inversion to preserve structural coherence during high-resolution image generation. Li et al.\cite{li2025inversion} propose Inversion-DPO, which circumvents reward modeling by reformulating Direct Preference Optimization (DPO) with DDIM inversion. Several studies confirm diffusion inversion's role in improving sampling quality. Bai et al\cite{bai2024zigzag} propose inserting inversion steps during sampling to improve output fidelity, while Zhou et al.\cite{Zhou_2025_Golden} show that using the initial noise derived from inversion as the starting point for denoising yields higher-quality images.

\section{Preliminaries}
In this section, we begin by introducing the preliminaries regarding diffusion models and Denoising Diffusion Implicit Model (DDIM), then describe the use of fixed-point iteration in diffusion inversion. 

\subsection{Diffusion Models}
Diffusion models~\cite{ddpm, song2020score} are a class of generative models that learn to generate data by reversing a gradual noising process. The framework consists of a forward process that progressively adds noise to data, and a reverse process that learns to denoise it.
The forward process is defined by gradually adding Gaussian noise to a data sample $\bm{z}_{t_0}$ over $N$ timesteps, producing a sequence of noisy latents $\bm{z}_{t_0}, \bm{z}_{t_1}, \dots, \bm{z}_{t_N}$:
\begin{equation}
  q(\bm{z}_{t_i}|\bm{z}_{t_{i-1}}) = \mathcal{N}(\bm{z}_{t_i}; \sqrt{\frac{\bar{\alpha}_{t_i}}{\bar{\alpha}_{t_{i-1}}}}\bm{z}_{t_{i-1}}, (1 - \frac{\bar{\alpha}_{t_i}}{\bar{\alpha}_{t_{i-1}}})\bm{I}),
\end{equation}
where $\{\bar\alpha_{t_i}\}_{i=1}^{N}$is the noise schedule. A notable property is that we can obtain the distribution of $\bm{z}_{t_i}$ at any timestep $t_i$ given $\bm{z}_{t_0}$:
\begin{equation}
  q(\bm{z}_{t_i} | \bm{z}_{t_0}) = \mathcal{N}(\bm{z}_{t_i}; \sqrt{\bar{\alpha}_{t_i}}\bm{z}_{t_0}, (1 - \bar{\alpha}_{t_i})\bm{I}).
\end{equation}
We can sample $\bm{z}_{t_i}$ from the distribution in a closed form:
 \begin{equation}
  \bm{z}_{t_i} = \sqrt{\bar{\alpha}_{t_i}}\bm{z}_{t_0} + \sqrt{1 - \bar{\alpha}_{t_i}}\bm{\epsilon}_{t_i},
  \label{predictedZ_t}
\end{equation}
where $\bm{\epsilon}_{t_i}$ is a random noise sampled from a Gaussian distribution $\mathcal{N}(\bm{0}, \bm{I})$.

The neural network is trained to estimate this noise by minimizing:
\begin{equation}
  \bm{\theta} = \mathop{\arg\min}\limits_{\bm{\theta}} \mathbb{E}_{t_i,\bm{z}_{t_0}, \bm{\epsilon}_{t_i}}[\lambda(t_i) \| \bm{\epsilon}_{\bm{\theta}}(\bm{z}_{t_i}, t_i, \bm{c}) - \bm{\epsilon} \|^2],
\end{equation}
where $\lambda(t_i)$ is a weighting function.

\subsection{DDIM Sampling and DDIM Inversion}
With the pre-trained model, we can generate images by denoising an initial Gaussian noise. The Denoising Diffusion Implicit Model (DDIM) \cite{ddim} sampler is a deterministic and more efficient alternative to the stochastic DDPM sampler for the denoising process. It leverages the same trained noise-prediction model $\bm{\epsilon}_{\bm{\theta}}(\cdot)$ but defines a non-Markovian forward process that allows for a deterministic denoising process.

Given $\bm{z}_{t_i}$, the denoising step of DDIM sampler to compute $\bm{z}_{t_{i-1}}$ is
\begin{equation}
  \begin{aligned}
    \bm{z}_{t_{i-1}} = 
    \sqrt{\bar{\alpha}_{t_{i-1}}}\bm{z}_{t_0}^{\bm{\theta}}(\bm{z}_{t_i}, t_i, \bm{c})
    + \sqrt{1-\bar{\alpha}_{t_{i-1}}} \bm{\epsilon}_{\theta}(\bm{z}_{t_{i}},t_{i}, \bm{c}),
  \end{aligned}
  \label{eq:ddim_sample}
\end{equation}
where $\bm{z}_{t_0}^{\bm{\theta}} (\bm{z}_{t_i}, t_i, \bm{c})$ is the data prediction at timestep $t_i$, that is using $\bm{z}_{t_i}$ to predict the $\bm{z}_{t_0}$ according to Eq.(~\ref{predictedZ_t}). It can be expressed in terms of the noise prediction $\bm{\epsilon}_{\theta}(\bm{z}_{t_{i}},t_{i}, \bm{c})$ as
\begin{equation}
  \bm{z}_{t_0}^{\bm{\theta}}(\bm{z}_{t_i}, t_i, \bm{c}) = \frac{\bm{z}_{t_{i}} - \sqrt{1-\bar{\alpha}_{t_{i}}}\bm{\epsilon}_{\theta}(\bm{z}_{t_{i}},t_{i}, \bm{c})}{\sqrt{\bar{\alpha}_{t_{i}}}}.
  \label{eq:data2noise}
\end{equation}

DDIM inversion is achieved by reversing the DDIM sampling equation (Eq. \eqref{eq:ddim_sample}), effectively running the deterministic generation process backwards. The inversion step from $\bm{z}_{t_{i-1}}$ to $\bm{z}_{t_i}$ is
\begin{equation}
  \begin{aligned}
    \bm{z}_{t_{i}} &= \frac{\sqrt{\bar{\alpha}_{t_{i}}}}{\sqrt{\bar{\alpha}_{t_{i-1}}}}\bm{z}_{t_{i-1}} \\
    &+ \left(\sqrt{1-\bar{\alpha}_{t_{i}}} - \frac{\sqrt{\bar{\alpha}_{t_{i}}}}{\sqrt{\bar{\alpha}_{t_{i-1}}}}\sqrt{1-\bar{\alpha}_{t_{i-1}}}\right) \bm{\epsilon}_{\theta}(\bm{z}_{t_{i}},t_{i}, \bm{c}).
  \end{aligned}
  \label{eq:ddim_inv}
\end{equation}
In an ideal, this equation is applied from $t_0$ to $t_N$ to obtain the corresponding initial noise $\bm{z}_{t_N}$. However, in practice, $\bm{\epsilon}_{\theta}(\bm{z}_{t_{i}},t_{i}, \bm{c})$ is approximated by $\bm{\epsilon}_{\theta}(\bm{z}_{t_{i-1}},t_{i-1}, \bm{c})$ since $\bm{z}_{t_{i}}$ is unknown. Due to the approximation, errors accumulate at each step, leading to imperfect reconstruction.

\subsection{Fixed-Point Iteration Method for DDIM Inversion}
Fixed-point iteration is a numerical method used to find a solution to equations of the form $x = g(x)$. The fixed point $x^{\ast}$ satisfies  
\begin{equation}
  x^{\ast} = g(x^{\ast}).
  \label{eq:fixed_point_converge}
\end{equation}

Since $\bm{\epsilon}_{\bm{\theta}}$ depends on $\bm{z}_{t_i}$, Eq. \eqref{eq:ddim_inv} can be expressed in the form of a fixed-point iteration:
\begin{equation}
  \begin{aligned}
    \bm{z}_{t_{i}}^{(n+1)} &= \bm{g}(\bm{z}_{t_{i}}^{(n)}), \quad n=0,1,2, \dots , \\
    \bm{g}(\bm{z}_{t_{i}}^{(n)}) &= \frac{\sqrt{\bar{\alpha}_{t_{i}}}}{\sqrt{\bar{\alpha}_{t_{i-1}}}}\bm{z}_{t_{i-1}}, \\
    &+ \left(\sqrt{1-\bar{\alpha}_{t_{i}}} - \frac{\sqrt{\bar{\alpha}_{t_{i}}}}{\sqrt{\bar{\alpha}_{t_{i-1}}}}\sqrt{1-\bar{\alpha}_{t_{i-1}}}\right) \bm{\epsilon}_{\theta}(\bm{z}_{t_{i}}^{(n)},t_{i}, \bm{c}).
  \end{aligned}
\end{equation}

In previous studies\cite{AIDI, garibi2024renoise},  the iteration of each inversion step is typically initialized with $\bm{z}_{t_{i-1}}$ set as the starting point $\bm{z}_{t_{i}}^{0}$. Under the local linearity assumption, the noise prediction at this point $\bm{z}_{t_{i-1}}$ should already be in proximity to that at the fixed point $\bm{z}_{t_{i}}$. However, empirical evidence from prior studies demonstrates that even when using this point as the initial state, fixed-point iteration still requires extensive iterations to achieve satisfactory reconstruction results.

\section{Our Method}
In this paper, we propose an iteration-free fixed-point estimator for diffusion inversion. We first derive an explicit expression for $\bm{z}_{t_{i}}$ without the noise neural network $\bm{\epsilon}_{\theta}(\bm{z}_{t_{i}},t_{i}, \bm{c})$. Then, we propose to estimate the $\bm{z}_{t_{i}}$ with error approximation. After that, we show how to perform initial estimation and give an analysis of our proposed fixed-point estimator.

\subsection{Explicit Expression of Fixed Point}
~\label{ExplicitExpression}
From Eq. \eqref{eq:ddim_inv} and Eq. \eqref{eq:fixed_point_converge}, it can be observed that the fixed point at each inversion step is $\bm{z}_{t_i}$. This motivates us to transform the implicit expression of the fixed point to an explicit expression. In Eq. \eqref{eq:ddim_inv}, $\bm{\epsilon}_{\theta}(\bm{z}_{t_{i}},t_{i}, \bm{c})$ is intractable due to the unknown input $\bm{z}_{t_i}$. To resolve this intractability, we leverage the interchangeable nature of data prediction and noise prediction (Eq. \eqref{eq:data2noise}). This allows us to decompose $\bm{\epsilon}_{\theta}(\bm{z}_{t_{i}},t_{i}, \bm{c})$ into a semi-unknown form:
\begin{equation}
  \bm{\epsilon}_{\theta}(\bm{z}_{t_{i}},t_{i}, \bm{c}) = \frac{\bm{z}_{t_i} - \sqrt{\bar{\alpha}_{t_i}}\bm{z}_{t_0}^{\bm{\theta}}(\bm{z}_{t_i}, t_i, \bm{c})}{\sqrt{1-\bar{\alpha}_{t_i}}},
\end{equation}
where $\bm{z}_{t_0}^{\bm{\theta}}(\bm{z}_{t_i}, t_i, \bm{c})$ is the data prediction at timestep $t_i$. Following ADM-ES\cite{bao2022analytic}, we assume the data prediction can be formulated as
\begin{equation}
  \bm{z}_{t_0}^{\bm{\theta}}(\bm{z}_{t_i}, t_i, \bm{c}) = \bm{z}_{t_0} + \bm{e}_{t_i},
\end{equation}
where $\bm{z}_{t_0}$ is the original image, which is known in diffusion inversion. And $\bm{e}_{t_i}$ represents the prediction error of the neural network at timestep $t_i$. 

With this reformulation, $\bm{z}_{t_i}$ is no longer an input to the network on the right-hand side of Eq. \eqref{eq:ddim_inv}, allowing it to be consolidated with the term $\bm{z}_{t_i}$ on the left-hand side of the equation. Hence, we can rewrite Eq. \eqref{eq:ddim_inv} as an explicit expression of the fixed point $\bm{z}_{t_{i}}$:
\begin{equation}
    \bm{z}_{t_{i}} = \frac{\sqrt{1-\bar{\alpha}_{t_{i}}}}{\sqrt{1-\bar{\alpha}_{t_{i-1}}}}\bm{z}_{t_{i-1}} + \eta_{t_i}\sqrt{\bar{\alpha}_{t_{i}}}(\bm{z}_{t_{0}} + \bm{e}_{t_{i}}),
  \label{eq:explicit_fixed_point}
\end{equation}
where
\begin{equation}
  \eta_{t_i} = \left(1 - \frac{\sqrt{\bar{\alpha}_{t_{i-1}}}}{\sqrt{\bar{\alpha}_{t_{i}}}}\frac{\sqrt{1-\bar{\alpha}_{t_{i}}}}{\sqrt{1-\bar{\alpha}_{t_{i-1}}}}\right).
\end{equation}

\subsection{Estimate Fixed Point with Error Approximation}
With $\bm{e}_{t_i}$ being the only remaining unknown in the explicit expression of the fixed point (Eq. \eqref{eq:explicit_fixed_point}), we propose the error approximation method to estimate the error in the data prediction at timestep $t_i$ by the one at preceding timestep $t_{i-1}$.
By rearranging Eq. \eqref{eq:explicit_fixed_point} at the preceding timestep $t_{i-1}$, we can obtain an expression for the data prediction error at that timestep:
\begin{equation}
    \bm{e}_{t_{i-1}} = 
    \frac{1}{\sqrt{\bar{\alpha}_{t_{i-2}}} \eta_{t_{i-1}}}\left(\bm{z}_{t_{i-2}} - \frac{\sqrt{1-\bar{\alpha}_{t_{i-2}}}}{\sqrt{1-\bar{\alpha}_{t_{i-1}}}}\bm{z}_{t_{i-1}} \right) 
    - \bm{z}_{t_{0}},
  \label{eq:err_approx_prev}
\end{equation}
where the values of $\bm{z}_{t_{i-1}}, \bm{z}_{t_{i-2}}$ and $\bm{z}_{t_0}$ have been known from the previous inversion steps. Thus, we can easily calculate $\bm{e}_{t_{i-1}}$ by this equation.
We propose to approximate the error in data prediction at the current inversion step by that at the preceding inversion step:
\begin{equation}
  \bm{e}_{t_i} \approx \bm{e}_{t_{i-1}}.
\end{equation}
We justify the rationality of this error approximation by providing both theoretical analysis and empirical evidence in Subsection \ref{ch:effective_analy}.
If $\bm{e}_{t_i}$ in Eq. \eqref{eq:explicit_fixed_point} is approximated by $\bm{e}_{t_{i-1}}$, the expression of the fixed point estimate $\hat{\bm{z}}_{t_{i}}$ is obtained:
\begin{equation}
  \hat{\bm{z}}_{t_{i}} = \frac{\sqrt{1-\bar{\alpha}_{t_{i}}}}{\sqrt{1-\bar{\alpha}_{t_{i-1}}}}\bm{z}_{t_{i-1}} + \eta_{t_i} \sqrt{\bar{\alpha}_{t_{i}}}(\bm{z}_{t_{0}} + \bm{e}_{t_{i-1}}),
  \label{eq:estimate_fixed_point}
\end{equation}
where $\bm{e}_{t_{i-1}}$ is calculated by Eq. \eqref{eq:err_approx_prev}. 

An accurate estimate of the fixed point provided by Eq. \eqref{eq:estimate_fixed_point} enables the inversion step to be completed in just one neural network evaluation without iterations:
\begin{equation}
  \begin{aligned}
    \bm{z}_{t_{i}} &= \frac{\sqrt{\bar{\alpha}_{t_{i}}}}{\sqrt{\bar{\alpha}_{t_{i-1}}}}\bm{z}_{t_{i-1}} \\
    &+ \left(\sqrt{1-\bar{\alpha}_{t_{i}}} - \frac{\sqrt{\bar{\alpha}_{t_{i}}}}{\sqrt{\bar{\alpha}_{t_{i-1}}}}\sqrt{1-\bar{\alpha}_{t_{i-1}}}\right) \bm{\epsilon}_{\theta}(\hat{\bm{z}}_{t_{i}}, t_{i}, \bm{c}).
  \end{aligned}
  \label{eq:ddim_inv_w_ours}
\end{equation}
Alg. \ref{alg:inv_process} outlines the overall inversion procedure incorporating our proposed method Iteration-Free Fixed-Point Estimator (IFE).

\begin{algorithm}[!t]
\renewcommand{\algorithmicrequire}{ \textbf{Input:}}     
\renewcommand{\algorithmicensure}{ \textbf{Output:}}    
\caption{Iteration-Free Fixed-Point Estimator for Diffusion Inversion.}
\begin{algorithmic}
\REQUIRE The original image $\bm{z}_{t_0}$  and its prompt $\bm{c}$. 
\STATE Perform the initial estimation with Eq. \eqref{eq:init_estimation}
\STATE Take the first inversion step with Eq. \eqref{eq:ddim_inv_w_ours}
\FOR{$i=2$ to $N$}
\STATE Employ error approximation with Eq. \eqref{eq:err_approx_prev}
\STATE Perform the estimation of the fixed point with Eq. \eqref{eq:estimate_fixed_point}
\STATE Take the inversion step with Eq. \eqref{eq:ddim_inv_w_ours}
\ENDFOR
\ENSURE The noise latent $\bm{z}_{t_N}$.
\end{algorithmic}
\label{alg:inv_process}
\end{algorithm}

\subsection{Initial Estimation}
Notably, at the first step of the inversion process, there is no previous inversion step; that is, there is no latent on the inversion trajectory prior to $\bm{z}_{t_0}$. Due to this absence, the error approximation is inapplicable at the first inversion step. 
Given the relatively high signal-to-noise ratio of the latent at $t_1$, the model's data prediction error is consequently small. This allows us to directly neglect the error term $\bm{e}_{t_1}$ in Eq. \eqref{eq:explicit_fixed_point},  thereby obtaining the expression for the initial fixed point estimate $\hat{\bm{z}}_{t_1}$ in the first inversion step.
\begin{equation}
  \hat{\bm{z}}_{t_{1}} = \frac{\sqrt{1-\bar{\alpha}_{t_{1}}}}{\sqrt{1-\bar{\alpha}_{t_{0}}}}\bm{z}_{t_{0}}
  + \eta_{t_1}\sqrt{\bar{\alpha}_{t_{1}}}\bm{z}_{t_{0}}. 
  \label{eq:init_estimation}
\end{equation}

\subsection{Analysis of Our Method} \label{ch:effective_analy}

\textit{Remark 1: Why can our iteration-free fixed-point estimator work?} First, we demonstrate that our method provides a better estimator than the direct use of $\bm{z}_{t_{i-1}}$ in DDIM inversion.

\begin{figure*}[!t]
    \centering
    \subfloat[]{\includegraphics[width=0.24\textwidth]{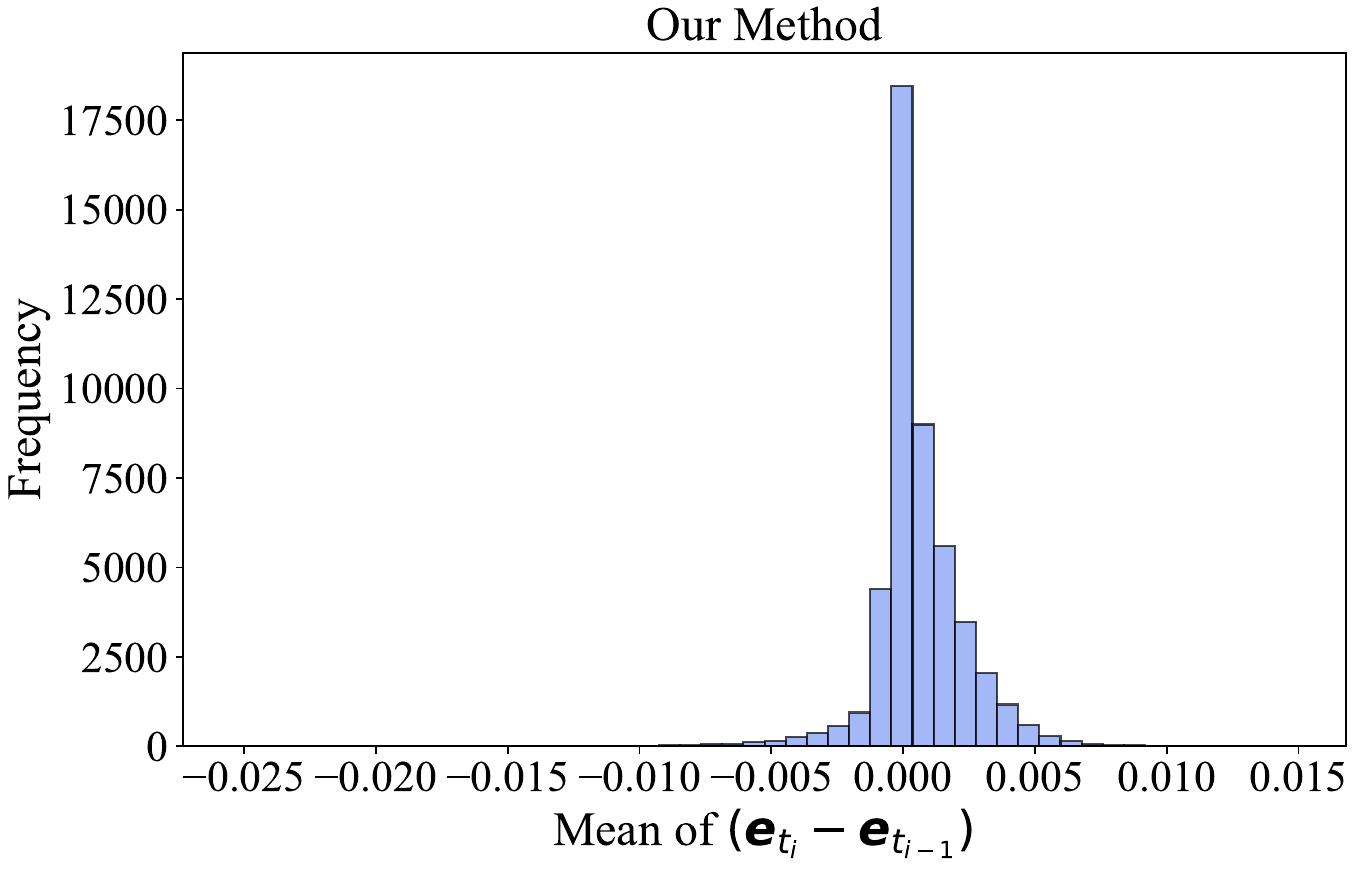}\label{fig:mean_diff_hist}}
    \hfill
    \subfloat[]{\includegraphics[width=0.24\textwidth]{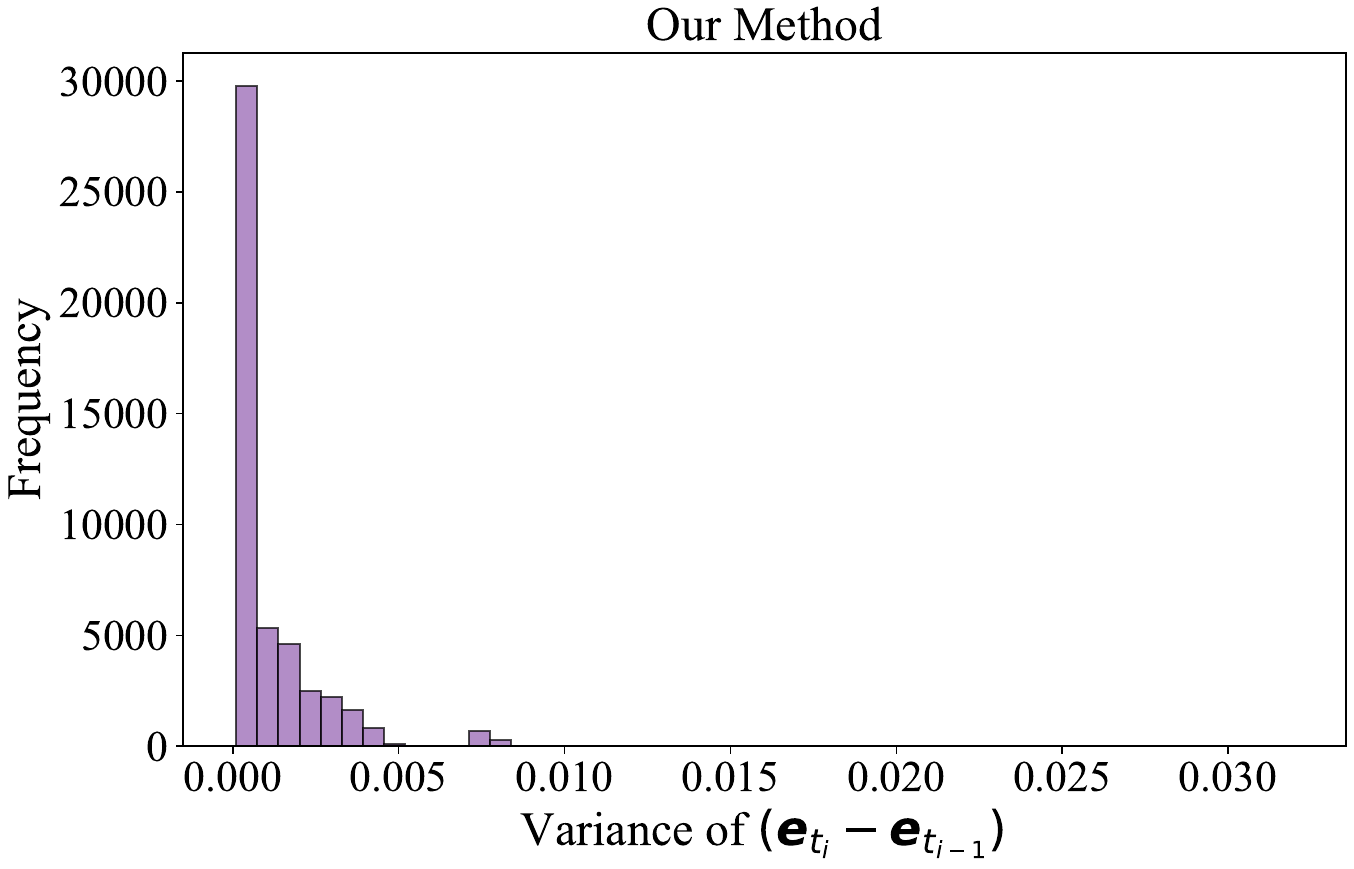}\label{fig:var_diff_hist}}
    \hfill
    \subfloat[]{\includegraphics[width=0.24\textwidth]{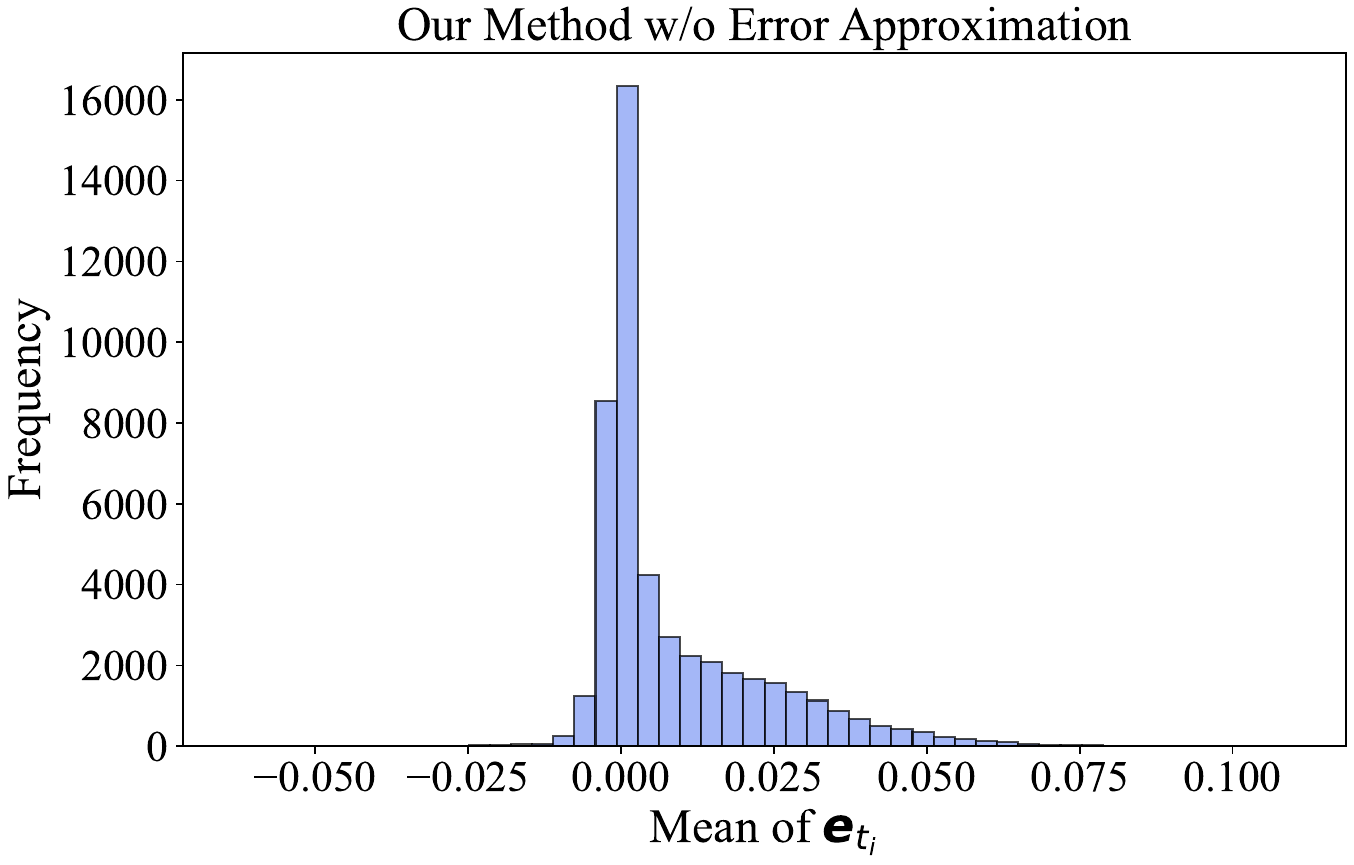}\label{fig:mean_e_hist}}
    \hfill
    \subfloat[]{\includegraphics[width=0.24\textwidth]{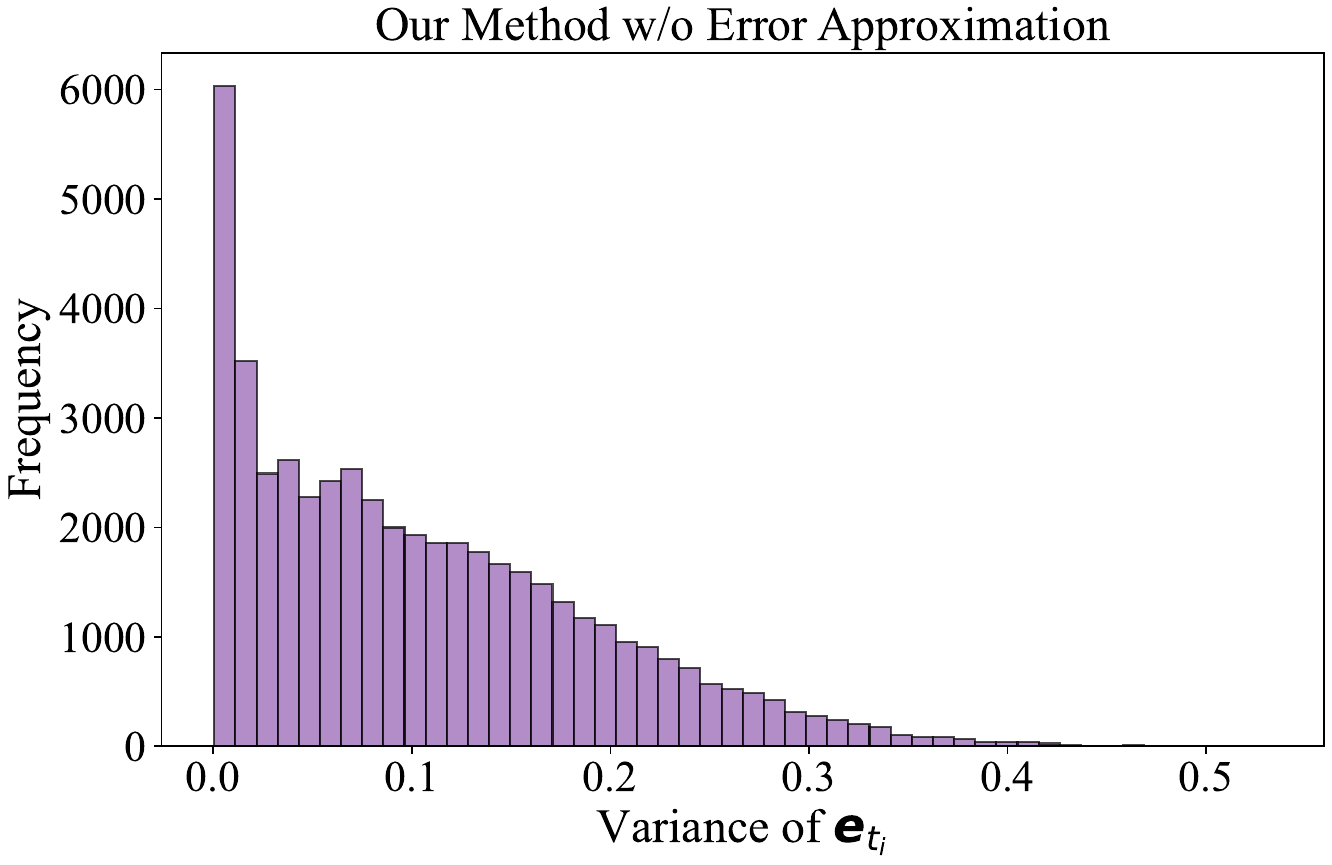}\label{fig:var_e_hist}}
    
    \caption{(a) The histogram of the sample mean of estimation error, which demonstrates that our method constitute an unbiased estimator of the fixed point (b) The histogram of the sample variance of estimation error, which demonstrates $\bm{e}_{t_i}$ and $\bm{e}_{t_{i-1}}$ are in proximity. (c) The histogram of the sample mean of estimation error when using our method without error approximation, which shifts from zero compared to our full method. (f) The histogram of the sample variance of estimation error when using our method without error approximation, whose overall variance is larger than that of our full method.}
    \label{fig:effective_statics_analy}
\end{figure*}

Let $\bm{z}_{t_i}$ denote the ground-truth fixed point, and let $\bm{z}$ denote an estimated latent of the fixed point (e.g., the estimates produced by our method or by DDIM inversion). The estimation error is defined as 
\begin{equation}
  \bm{\Delta}_{t_{i}}(\bm{z}) = \bm{z}_{t_i} - \bm{z}.
\end{equation}
Following ADM-ES\cite{ning2024admes}, we model the error in data prediction by a Gaussian distribution:
\begin{equation}
  \bm{e}_{t_i} \sim \mathcal{N}(\bm{0}, \gamma_{t_i}\bm{I}),
\end{equation}
where $\gamma_{t_i}$ is a time-dependent scalar.

In DDIM inversion, we can calculate the estimation error by taking $\bm{z}_{t_{i-1}}$ as the input latent:
\begin{equation}
  \begin{aligned}
    \bm{\Delta}_{t_{i}}(\bm{z}_{t_{i-1}}) &= \bm{z}_{t_i} - \bm{z}_{t_{i-1}} \\
    &= \left(\frac{\sqrt{1-\bar{\alpha}_{t_{i}}}}{\sqrt{1-\bar{\alpha}_{t_{i-1}}}} - 1\right) \bm{z}_{t_{i-1}} + \eta_{t_i}\sqrt{\bar{\alpha}_{t_{i}}}(\bm{z}_{t_{0}} + \bm{e}_{t_{i}}) \\
    &\sim \mathcal{N}\left(\bm{\mu}_{t_i}, \eta_{t_i}^2 \bar{\alpha}_{t_{i}} \gamma_{t_i}\bm{I}\right),
  \end{aligned}
\end{equation}
where the mean of $\bm{\Delta}_{t_{i}}(\bm{z}_{t_{i-1}})$ can be calculated by
\begin{equation}
  \bm{\mu}_{t_i} = \left(\frac{\sqrt{1-\bar{\alpha}_{t_{i}}}}{\sqrt{1-\bar{\alpha}_{t_{i-1}}}} - 1\right) \bm{z}_{t_{i-1}} + \eta_{t_i}\sqrt{\bar{\alpha}_{t_{i}}}\bm{z}_{t_{0}}.
\end{equation}
Since the mean is non-zero, the approximate expression of the fixed point in DDIM inversion constitutes a \textbf{biased} estimator.

In our method, we use $\hat{\bm{z}}_{t_{1}}$ expressed in Eq. \eqref{eq:estimate_fixed_point} to estimate the fixed point, which produces a smaller estimation error:
\begin{equation}
  \begin{aligned}
    \bm{\Delta}_{t_{i}}(\hat{\bm{z}}_{t_{i}}) &= \bm{z}_{t_i} - \hat{\bm{z}}_{t_{i}} \\
    &= \eta_{t_i}\sqrt{\bar{\alpha}_{t_{i}}}(\bm{e}_{t_{i}} - \bm{e}_{t_{i-1}}) \\
    &\sim \mathcal{N}\left(\bm{0}, \eta_{t_i}^2 \bar{\alpha}_{t_{i}} (\gamma_{t_i} + \gamma_{t_{i-1}})\bm{I} - 2\mathrm{Cov}(\bm{e}_{t_{i}}, \bm{e}_{t_{i-1}})\right).
  \end{aligned}
\end{equation}
Due to the zero mean, our method yields an \textbf{unbiased} estimate of the fixed point. This analysis shows that our estimation is better than the DDIM inversion. In our experiments, we further show that our iteration-free estimator achieves reconstruction performance comparable to that of fixed-point-iteration-based methods, yet with significantly faster inference.

\textit{Remark 2: Efficiency of the error approximation.} In our method, we use the preceding estimation error as the approximation of the unknown one. To evaluate its efficiency, we propose a variant of our method without the error approximation, which is formulated as
 \begin{equation}
  \tilde{\bm{z}}_{t_{i}} = \frac{\sqrt{1-\bar{\alpha}_{t_{i}}}}{\sqrt{1-\bar{\alpha}_{t_{i-1}}}}\bm{z}_{t_{i-1}}
  + \eta_{t_i}\sqrt{\bar{\alpha}_{t_{i}}}\bm{z}_{t_{0}}
  \label{eq:ours_wo_err_approx}
\end{equation}


The fixed point estimation error of this baseline is
\begin{equation}
  \begin{aligned}
    \bm{\Delta}_{t_{i}}(\tilde{\bm{z}}_{t_{i}}) &= \bm{z}_{t_i} - \tilde{\bm{z}}_{t_{i}} \\
    &= \eta_{t_i}\sqrt{\bar{\alpha}_{t_{i}}}\bm{e}_{t_{i}} \\
    &\sim \mathcal{N}\left(\bm{0}, \eta_{t_i}^2 \bar{\alpha}_{t_{i}}\gamma_{t_i}\bm{I}\right).
  \end{aligned}
\end{equation}
While the variant of our method without error approximation also produces a zero-mean estimation error, it alters the error variance. Based on the mathematical form of this variance, it is difficult to determine whether this variance is larger than the estimation error variance of our full method. We conduct an experiment to verify it. In this empirical validation, we generate 1,000 random images using Stable Diffusion XL~\cite{sdxl} and record the data prediction error $\bm{e}_{t_i}$ from the first 49 denoising steps for statistical analysis. The total number of denoising steps is 50.

As shown in Fig. \ref{fig:mean_diff_hist}, the overall mean of $\bm{e}_{t_i} - \bm{e}_{t_{i-1}}$ is heavily concentrated at and around zero, indicating our method constitutes an unbiased estimator for the fixed point. Fig. \ref{fig:var_diff_hist} depicts the sample variance of $\bm{e}_{t_i} - \bm{e}_{t_{i-1}}$, which is equivalent to its mean squared error (MSE) from the zero matrix. It can be observed that the MSE for the majority of samples is only marginally greater than zero, indicating that $\bm{e}_{t_i}$ and $\bm{e}_{t_{i-1}}$ are approximate and, consequently, that our error variance is small.
 
Fig. \ref{fig:var_e_hist} presents the histogram of error variance for the variant without error approximation. Compared to the variance histogram of our full method in Fig. \ref{fig:var_diff_hist}, the variance of the variant is shifted towards larger values overall, indicating a higher MSE relative to the zero matrix. In addition, Fig. \ref{fig:mean_e_hist} depicts the histogram of the error mean for the variant without error approximation. It shows that the mean values still predominantly cluster near zero. However, compared to the histogram for our full method in Fig. \ref{fig:mean_diff_hist}, the frequency of mean values deviating from zero is noticeably higher. In summary,  we empirically verified that our full method achieves a \textbf{lower estimation error variance} than the variant without error approximation.

\section{Experiments}
We present a comprehensive empirical analysis of our method. First, subsection \ref{ch:ds_settings} outlines our experimental setup. Subsection \ref{ch:comparison} demonstrates our approach's superior performance over existing techniques in both quantitative metrics and visual quality. Subsection \ref{ch:ours_vs_fpi} validates that our method matches multi-iteration baseline performance without costly extra iterations. 
Finally, subsection \ref{ch:ablation} verifies the contribution of each component. Together, these sections provide rigorous evidence for our method's effectiveness and efficiency.

\subsection{Datasets and Settings} \label{ch:ds_settings}
We implement reconstruction experiments on two image-caption datasets: MS-COCO 2017\cite{MSCOCO} and NOCAPS\cite{NOCAPS}. 

\textbf{MS-COCO 2017 Validation Set:} The MS-COCO (Microsoft Common Objects in Context) 2017 validation set is a standard and densely annotated benchmark in computer vision. It contains 5,000 images encompassing 80 common object categories in complex, real-world contexts. Due to its high-quality annotations and the prevalence of its training set in model pre-training, it serves as the primary benchmark for evaluating in-domain performance and comparative accuracy. We extract 2,298 images from the MS-COCO 2017 validation and test sets\cite{MSCOCO} for the comparative experiment, using the extraction script from EasyInv.

\textbf{NOCAPS Validation Set:} The NOCAPS (Novel Object Captioning at Scale) validation set is designed to evaluate model performance in an open-world setting. It consists of images sourced from the Open Images dataset, containing many object categories that are not present in the training set of common models like MS-COCO. 
We extract 4,498 images from the validation set with both height and width not exceeding 1,024 for all experiments. Unless otherwise specified, all image reconstructions employ the first caption associated with each image in the dataset as the default text prompt.

All experiments are conducted on an NVIDIA RTX 3090 GPU using pre-trained models, Stable Diffusion v1.4\cite{Rombach_2022_CVPR} and Stable Diffusion XL\cite{sdxl}. The guidance scales of both the inversion and denoising processes are set to 1.0. 

To demonstrate the superiority of our method in terms of both performance and efficiency, four metrics are employed for evaluation. PSNR, LPIPS\cite{LPIPS}, SSIM\cite{SSIM}, and CLIP-I are used to assess reconstruction faithfulness, and NFE (Number of Function Evaluations) is used to evaluate inversion efficiency. CLIP-I is the average cosine similarity of the CLIP\cite{CLIP} embeddings between all original and reconstructed image pairs.

\subsection{Comparison with Prior Methods} \label{ch:comparison}
We compare our method against DDIM Inversion, ReNoise~\cite{garibi2024renoise}, AIDI~\cite{AIDI}, and EasyInv~\cite{zhang2024easyinv}. In these baselines, AIDI and ReNoise employ fixed-point iteration, whereas EasyInv is a recent state-of-the-art method. This experimental comparison provides a clearer and more comprehensive evaluation of our method’s advantages.

\begin{table}[!t]
    \belowrulesep=0pt
    \aboverulesep=0pt
    \renewcommand{\arraystretch}{1.2}
    \caption{Comparative result of image reconstruction on MS-COCO 2017}
    \centering
    \resizebox{1\columnwidth}{!}
    {
    \begin{tabular}{| c | c | c | c | c |} 
        \toprule
        & LPIPS ($\downarrow$) & SSIM ($\uparrow$) & PSNR ($\uparrow$) & NFE ($\downarrow$) \\ [0.5ex] 
        \midrule
        DDIM Inversion & 0.328  & 0.621 & 29.717 & \textbf{50}\\
        ReNoise & 0.316 & 0.641 & 31.025 & 150\\
        AIDI & 0.373 & 0.563 & 29.107 & 140\\
        EasyInv & 0.321 & 0.646 & 30.189 & \textbf{50} \\
        IFE (Ours) & \textbf{0.216} & \textbf{0.733} & \textbf{31.658} & \textbf{50} \\
        \bottomrule
    \end{tabular}
    }
    \label{tbl:comparison_coco}
\end{table}

\begin{table}[!t]
    \belowrulesep=0pt
    \aboverulesep=0pt
    \renewcommand{\arraystretch}{1.2}
    \caption{Comparative result of image reconstruction on NOCAPS}
    \centering
    \resizebox{1\columnwidth}{!}
    {
    \begin{tabular}{| c | c | c | c | c | c |} 
        \toprule
        & LPIPS ($\downarrow$) & SSIM ($\uparrow$) & PSNR ($\uparrow$) & CLIP-I ($\uparrow$) & NFE ($\downarrow$) \\ [0.5ex] 
        \midrule
        DDIM Inversion & 0.314  & 0.777 & 31.190 & 0.901 & \textbf{50}\\
        ReNoise & 0.241 & 0.786 & 33.067 & 0.933 & 150\\
        AIDI & 0.221 & 0.837 & 33.153 & 0.960 & 150\\
        EasyInv & 0.288 & 0.678 & 28.494 & 0.951 & \textbf{50} \\
        IFE (Ours) & \textbf{0.205} & \textbf{0.849} & \textbf{33.484} & \textbf{0.968} & \textbf{50} \\
        \bottomrule
    \end{tabular}
    }
    \label{tbl:comparison_nocaps}
\end{table}

Table \ref{tbl:comparison_coco} shows a comparative result of image reconstruction on the MS-COCO 2017 dataset. In this experiment, our setup follows EasyInv: we use Stable Diffusion v1.4 and set the number of inversion and denoising steps to 50 for all methods, with the exception of AIDI\cite{AIDI}, which is recommended to set steps to 20. All results of Table \ref{tbl:comparison_coco}, except ours, are derived directly from the EasyInv paper, while the NFE values are sourced from the respective official publications or codebases of the compared methods. Our method achieves superior performance on the LPIPS, PSNR, and SSIM metrics while maintaining the lowest NFE. Compared to EasyInv, our method yields a 32.7\% reduction in LPIPS and improvements of 13.5\% in SSIM and 4.9\% in PSNR with a comparable NFE.

We further expand our evaluation to the NOCAPS validation dataset using Stable Diffusion XL for a more comprehensive analysis. This dataset provides images with higher resolutions than MS-COCO 2017, thereby offering a more robust validation of our method's effectiveness and generalizability. We set the number of inversion and denoising steps to 50 for all methods. The NFE per step is set to 1 for all methods, with the exception of ReNoise and AIDI, which use an NFE of 3 per step. Since the official implementation of AIDI is unavailable, we re-implement the method ourselves based on its algorithmic description. The superiority of our method is evidenced by the results in Table \ref{tbl:comparison_nocaps}, which show leading scores on the LPIPS, SSIM, PSNR, and CLIP-I metrics. Compared to ReNoise, our method yields a 15.1\% reduction in LPIPS and improvements of 8.0\% in SSIM, 1.3\% in PSNR, and 3.8\% in CLIP-I with a comparable NFE.

\begin{figure*}[!t]
    \centering
    \includegraphics[width=1.0\textwidth]{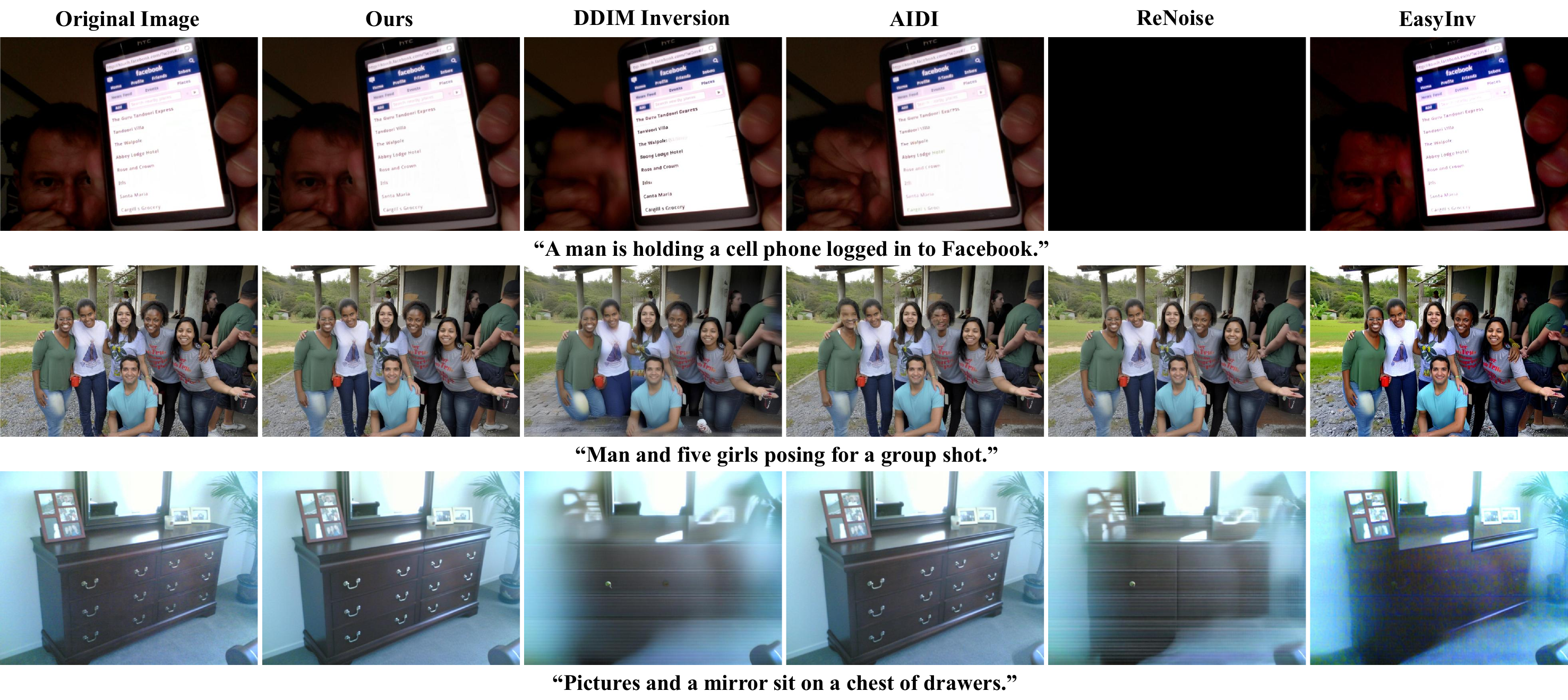}
    \caption{Qualitative Comparison: Ours vs. DDIM Inversion, AIDI, ReNoise, and EasyInv.}
    \label{fig:comparative_expr}
\end{figure*}

To visually validate the quantitative findings, we present representative examples from the NOCAPS validation dataset in Fig. \ref{fig:comparative_expr}. The qualitative comparisons reveal distinct advantages of our method in reconstruction accuracy. As shown in the first row of Fig. \ref{fig:comparative_expr}, when reconstructing images with high dynamic range, our method demonstrates superior capability. Both DDIM inversion and AIDI are observed to lose critical facial features. Conversely, ReNoise produces a completely black image due to gradient instability during its iterative optimization, whereas EasyInv suffers from over-saturation, resulting in the loss of high-frequency details in the dark regions of the original image. Turning to the reconstruction results in the second row of this figure, which features a group photograph with a cluttered background, several distinct failure modes are observed among the compared methods. DDIM inversion and AIDI produce significant artifacts, while ReNoise loses certain leg details, and EasyInv appears over-saturated. In contrast, our method exhibits only a minor loss of detail, confined specifically to the hand of the woman in green on the far left. In the third row of this figure, the reconstructions from DDIM inversion and ReNoise exhibit blurring and artifacts, while AIDI's result deviates from the original image in the fine details of the bottom-right corner. Meanwhile, EasyInv's output remains oversaturated and suffers from blur. In contrast, our method successfully avoids all of these issues.

\subsection{Comparison with Multi-Iteration Baselines} \label{ch:ours_vs_fpi}

\begin{table}[t]
\belowrulesep=0pt
\aboverulesep=0pt
\renewcommand{\arraystretch}{1.2}
\caption{Comparative reconstruction experiment of inversion efficiency}
\centering
\begin{tabular}{@{}|c|c|c|c|c|c|@{}}
\toprule
 &
  \begin{tabular}[c]{@{}c@{}}Additional \\ Iteration Times \\ Per Step\end{tabular} &
  LPIPS &
  SSIM &
  PSNR &
  CLIP-I \\ \midrule
\multirow{4}{*}{\begin{tabular}[c]{@{}c@{}}Baseline\end{tabular}} &
  1 &
  0.2254 &
  0.8345 &
  33.061 &
  0.957 \\
     & 2 & 0.2079 & 0.8440 & 33.440 & 0.964 \\
     & 3 & 0.1985 & 0.8500 & 33.656 & 0.970 \\
     & 4 & 0.2006 & 0.8484 & 33.608 & 0.969 \\ \midrule
IFE (Ours) & 0 & 0.2047 & 0.8486 & 33.484 & 0.968 \\ \bottomrule
\end{tabular}
\label{tbl:inv_efficiency}
\end{table}

We conduct a comparative study in order to demonstrate that our proposed non-iterative inversion method achieves results comparable to those produced by the fixed-point-based method with multiple iterations, while significantly reducing computational overhead during the inversion process. To be specific, the fixed-point-based inversion method, without any acceleration techniques (e.g., AIDI\cite{AIDI} or ReNoise\cite{garibi2024renoise}), is used as the baseline against which our method is compared. We use Stable Diffusion XL on the NOCAPS validation dataset to reconstruct images. For both our method and the baseline, the number of inversion and denoising steps is fixed at 50. During the inversion process, our method performs one basic iteration per step (i.e., NFE=1 per step). In contrast, the baseline conducts one to four additional iterations beyond the basic one per step, resulting in an NFE range of 2 to 5 per step. Our method delivers performance superior to fixed-point iteration at 2 additional iterations per step and remains competitive with the 3-additional-iteration configuration, despite requiring fewer iterations, as shown in Table \ref{tbl:inv_efficiency}. This result implies that our method enables us to arrive at a point near that of the fixed-point method at 2 additional iterations while avoiding superfluous iterations. Furthermore, the results in the table reveal a non-monotonic relationship between the number of iterations and baseline performance. Under our experimental configuration, the baseline achieves its optimal performance with three additional iterations. This non-monotonicity suggests potential convergence instability of the iterative process, which consequently complicates the selection of an optimal iteration count. In contrast, our method achieves near-optimal performance directly, eliminating the need for tedious hyperparameter selection.

\begin{figure*}[!t]
    \centering
    \includegraphics[width=1.0\textwidth]{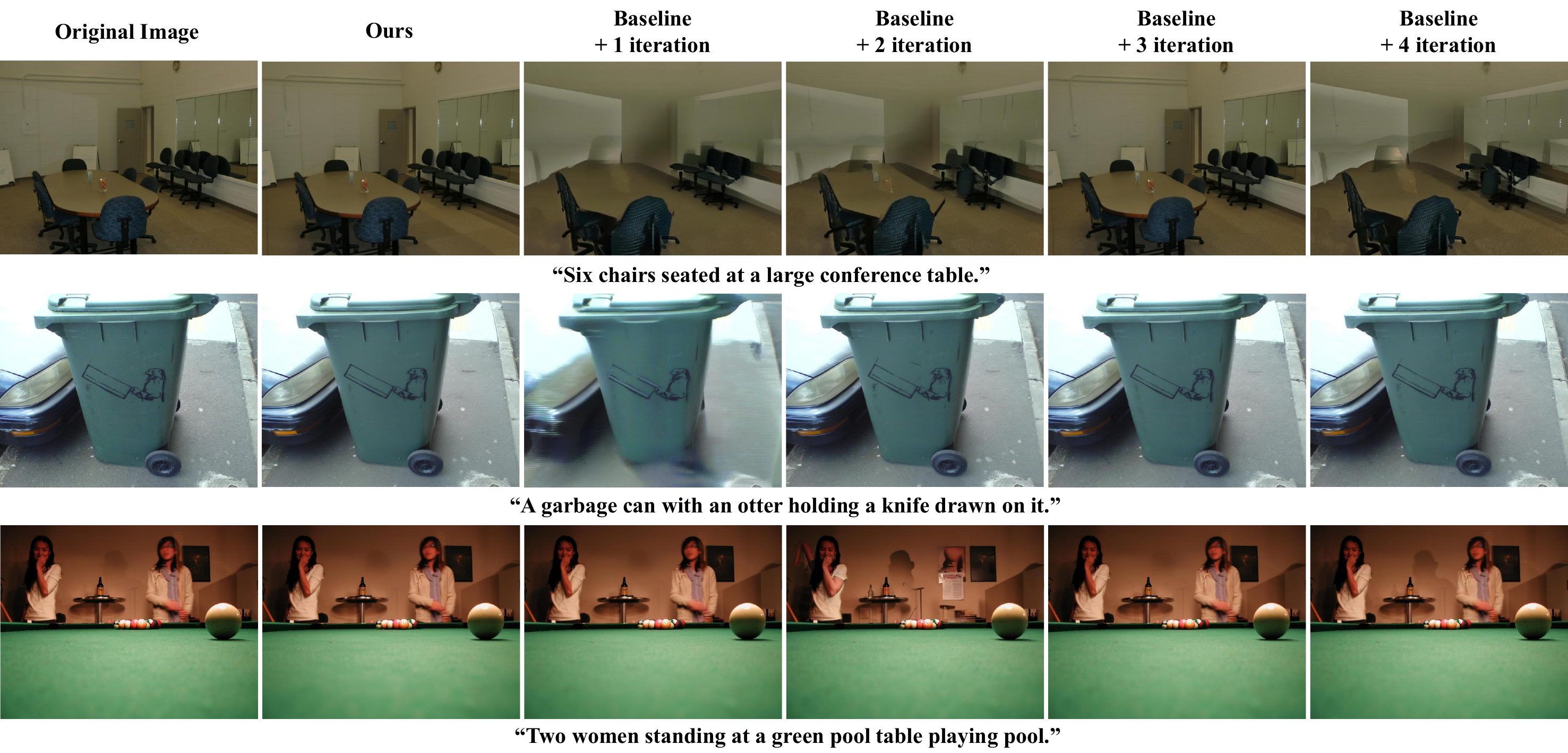}
    \caption{Visual Examples: Ours vs. Iteration-Sensitive Baselines}
    \label{fig:ours_vs_fpi}
\end{figure*}

Although the quantitative comparisons present a compelling case for our method, a qualitative visual comparison is indispensable for a comprehensive understanding. We provide visual examples from the NOCAPS validation dataset to demonstrate that our method achieves a consistent near-optimal result. As shown in Fig. \ref{fig:ours_vs_fpi}, our method consistently achieves a reconstruction quality that closely matches the peak performance of the baseline. It is noteworthy that whereas our approach offers stable performance without the need for hyperparameter tuning, the reconstruction accuracy of the baseline is highly dependent on the number of additional iterations per step. A configuration of additional iterations that is effective for one image may prove ineffective for another. For instance, in the second row of Fig. \ref{fig:ours_vs_fpi}, the baseline with one additional iteration exhibits poor quality, yet it yields a high-quality reconstruction with two additional iterations; the third row, however, shows the completely divergent behavior.

\subsection{Ablation Study} \label{ch:ablation}
To validate the effectiveness of the two components of our method: error approximation and initial estimation, we conduct a comprehensive ablation study. We compare our full method against two ablated variants: (A) the method without the error approximation, which directly neglects the error term $\bm{e}_{t_{i}}$ and estimates the fixed point with Eq. \eqref{eq:ours_wo_err_approx}; (B) the method starting inversion without initial estimation at the first inversion step. All methods are evaluated identically on the NOCAPS validation dataset using Stable Diffusion XL. The inversion and denoising steps are both fixed at 50. 

\begin{table}[!t]
    \belowrulesep=0pt
    \aboverulesep=0pt
    \renewcommand{\arraystretch}{1.2}
    \caption{Quantitative ablation study on NOCAPS validation dataset.}
    \centering
    \resizebox{1\columnwidth}{!}
    {
    \begin{tabular}{| c | c | c | c | c |} 
        \toprule
        & LPIPS ($\downarrow$) & SSIM ($\uparrow$) & PSNR ($\uparrow$) & CLIP-I ($\uparrow$) \\ [0.5ex] 
        \midrule
        Ours & \textbf{0.2047} & \textbf{0.8486} & \textbf{33.484} & \textbf{0.9682}\\
        Ours w/o error approximation & 0.2769 & 0.8050 & 31.766 & 0.9400\\
        Ours w/o initial estimation & 0.2093 & 0.8472 & 33.442 & 0.9668\\
        \bottomrule
    \end{tabular}
    }
    \label{tbl:ablation}
\end{table}

The results detailed in Table \ref{tbl:ablation} present the effectiveness of each component in our method. First, the removal of the error approximation (Variant A) leads to a catastrophic performance drop, with a significant rise in LPIPS (from 0.2047 to 0.2769) and a sharp decline in SSIM (from 0.8486 to 0.8050), PSNR(from 33.484 to 31.766), and CLIP-I (from 0.9682 to 0.9400). This unequivocally establishes that the error approximation is the key component responsible for achieving high accuracy of diffusion inversion. Second, the absence of initial estimation slightly degrades the performance, with an increase in LPIPS (from 0.2047 to 0.2093) and a decrease in SSIM (from 0.8486 to 0.8472), PSNR(from 33.484 to 33.442), and CLIP-I (from 0.9682 to 0.9668). Given these results, the application of the initial estimation remains advisable, as it provides a modest improvement in performance without introducing any appreciable computational cost.

\section{Conclusion and Future Works}
In this work, we introduced Iteration-Free Fixed-Point Estimator, a novel training-free diffusion inversion method designed to address the computational inefficiency of fixed-point iteration methods. Our theoretical contribution, the explicit estimate expression of the fixed point in the inversion step, provides an unbiased estimator of the fixed point with a low variance for the neural network input. Practically, this insight allows for the development of highly efficient inversion methods, making inversion-based downstream tasks like style transfer and object manipulation more accessible by eliminating the need for costly iterative refinements. 
Through extensive evaluation on the NOCAPS and MS-COCO datasets, we demonstrated that our method achieves superior image reconstruction fidelity while significantly reducing computational overhead. We acknowledge that our study is primarily focused on the DDIM sampler. The performance of our method under more complex or stochastic sampling schemes remains an open issue to be solved.

\bibliography{references.bib}

@inproceedings{ddpm,
 author = {Ho, Jonathan and Jain, Ajay and Abbeel, Pieter},
 booktitle = {Adv. Neural Inform. Process. Syst.},
 pages = {6840--6851},
 title = {Denoising Diffusion Probabilistic Models},
 volume = {33},
 year = {2020}
}

@article{ClassifierG,
  title={Diffusion Models Beat {GAN}s on Image Synthesis},
  author={Dhariwal, Prafulla and Nichol, Alexander},
  journal={Adv. Neural Inform. Process. Syst.},
  volume={34},
  pages={8780--8794},
  year={2021}
}

@article{CFG,
  title={Classifier-free Diffusion Guidance},
  author={Ho, Jonathan and Salimans, Tim},
  journal={arXiv preprint arXiv:2207.12598},
  year={2021}
}

@InProceedings{iddpm,
  title = 	 {Improved Denoising Diffusion Probabilistic Models},
  author =       {Nichol, Alexander Quinn and Dhariwal, Prafulla},
  booktitle = 	 {Int. Conf. Mach. Learn.},
  pages = 	 {8162--8171},
  year = 	 {2021},
  volume = 	 {139},
}

@inproceedings{ldm,
  title={High-Resolution Image Synthesis with Latent Diffusion Models},
  author={Rombach, Robin and Blattmann, Andreas and Lorenz, Dominik and Esser, Patrick and Ommer, Bj{\"o}rn},
  booktitle={IEEE Conf. Comput. Vis. Pattern Recog.},
  pages={10684--10695},
  year={2022}
}

@article{bao2022analytic,
  title={Analytic-dpm: an analytic estimate of the optimal reverse variance in diffusion probabilistic models},
  author={Bao, Fan and Li, Chongxuan and Zhu, Jun and Zhang, Bo},
  journal={arXiv preprint arXiv:2201.06503},
  year={2022}
}

@ARTICLE{tmm11207541,
  author={Li, Yinqi and Chang, Hong and Hou, Ruibing and Shan, Shiguang and Chen, Xilin},
  journal={IEEE Transactions on Multimedia}, 
  title={{DIVE}: Inverting Conditional Diffusion Models for Discriminative Tasks}, 
  year={2025},
  volume={},
  number={},
  pages={1-12},
  note={early access, DOI: {
\href{http://doi.org/10.1109/TMM.2025.3623508}{10.1109/TMM.2025.3623508}}},
  doi={10.1109/TMM.2025.3623508}}

@ARTICLE{tmm11153996,
  author={Taghipour, Ashkan and Ghahremani, Morteza and Bennamoun, Mohammed and Rekavandi, Aref Miri and Laga, Hamid and Boussaid, Farid},
  journal={IEEE Transactions on Multimedia}, 
  title={Box It to Bind It: Unified Layout Control and Attribute Binding in Text-to-Image Diffusion Models}, 
  year={2025},
  volume={27},
  number={},
  pages={8393-8407},
}

@ARTICLE{tmm11071375,
  author={Chen, Mulin and Wang, Yajie and Li, Xuelong},
  journal={IEEE Transactions on Multimedia}, 
  title={{PIMG}: Progressive Image-to-Music Generation With Contrastive Diffusion Models}, 
  year={2025},
  volume={27},
  number={},
  pages={6732-6739},
}

@inproceedings{hertz2022prompt,
  title={Prompt-to-prompt image editing with cross attention control},
  author={Hertz, Amir and Mokady, Ron and Tenenbaum, Jay and Aberman, Kfir and Pritch, Yael and Cohen-Or, Daniel},
  booktitle={Int. Conf. Learn. Represent.},
  year={2023}
}

@inproceedings{tumanyan2023plug,
  title={Plug-and-play diffusion features for text-driven image-to-image translation},
  author={Tumanyan, Narek and Geyer, Michal and Bagon, Shai and Dekel, Tali},
  booktitle={IEEE Conf. Comput. Vis. Pattern Recog.},
  pages={1921--1930},
  year={2023}
}

@inproceedings{chung2024style,
  title={Style injection in diffusion: A training-free approach for adapting large-scale diffusion models for style transfer},
  author={Chung, Jiwoo and Hyun, Sangeek and Heo, Jae-Pil},
  booktitle={IEEE Conf. Comput. Vis. Pattern Recog.},
  pages={8795--8805},
  year={2024}
}

@article{vontobel2025hiwave,
  title={HiWave: Training-Free High-Resolution Image Generation via Wavelet-Based Diffusion Sampling},
  author={Vontobel, Tobias and Sadat, Seyedmorteza and Salehi, Farnood and Weber, Romann M},
  journal={arXiv preprint arXiv:2506.20452},
  year={2025}
}

@inproceedings{li2025inversion,
  title={Inversion-DPO: Precise and Efficient Post-Training for Diffusion Models},
  author={Li, Zejian and Li, Yize and Meng, Chenye and Liu, Zhongni and Ling, Yang and Zhang, Shengyuan and Yang, Guang and Yang, Changyuan and Yang, Zhiyuan and Sun, Lingyun},
  booktitle={ACM Int. Conf. Multimedia},
  pages = {9901–9910},
  year={2025}
}

@inproceedings{ddim,
  title={Denoising diffusion implicit models},
  author={Song, Jiaming and Meng, Chenlin and Ermon, Stefano},
  booktitle={Int. Conf. Learn. Represent.},
  year={2021}
}

@inproceedings{AIDI,
  title={Effective real image editing with accelerated iterative diffusion inversion},
  author={Pan, Zhihong and Gherardi, Riccardo and Xie, Xiufeng and Huang, Stephen},
  booktitle={Int. Conf. Comput. Vis.},
  pages={15912--15921},
  year={2023}
}

@inproceedings{garibi2024renoise,
  title={{ReNoise}: Real image inversion through iterative noising},
  author={Garibi, Daniel and Patashnik, Or and Voynov, Andrey and Averbuch-Elor, Hadar and Cohen-Or, Daniel},
  booktitle={Eur. Conf. Comput. Vis.},
  pages={395--413},
  year={2024},
}

@InProceedings{zhang2024easyinv,
  title = 	 {{E}asy{I}nv: Toward Fast and Better {DDIM} Inversion},
  author =       {Zhang, Ziyue and Lin, Mingbao and Yan, Shuicheng and Ji, Rongrong},
  booktitle = 	 {Int. Conf. Mach. Learn.},
  pages = 	 {75503--75512},
  year = 	 {2025},
  volume = 	 {267},
}

@inproceedings{mokady2023null,
  title={Null-text inversion for editing real images using guided diffusion models},
  author={Mokady, Ron and Hertz, Amir and Aberman, Kfir and Pritch, Yael and Cohen-Or, Daniel},
  booktitle={IEEE Conf. Comput. Vis. Pattern Recog.},
  pages={6038--6047},
  year={2023}
}

@inproceedings{dong2023prompt,
  title={Prompt tuning inversion for text-driven image editing using diffusion models},
  author={Dong, Wenkai and Xue, Song and Duan, Xiaoyue and Han, Shumin},
  booktitle={Int. Conf. Comput. Vis.},
  pages={7430--7440},
  year={2023}
}

@inproceedings{jupnp,
  title={{PnP} Inversion: Boosting Diffusion-based Editing with 3 Lines of Code},
  author={Ju, Xuan and Zeng, Ailing and Bian, Yuxuan and Liu, Shaoteng and Xu, Qiang},
  booktitle={Int. Conf. Learn. Represent.},
  year={2024}
}

@inproceedings{ning2024admes,
  title={Elucidating the Exposure Bias in Diffusion Models},
  author={Ning, Mang and Li, Mingxiao and Su, Jianlin and Salah, Albert Ali and Ertugrul, Itir Onal},
  booktitle={Int. Conf. Learn. Represent.},
  year={2024}
}

@article{karras2022elucidating,
  title={Elucidating the design space of diffusion-based generative models},
  author={Karras, Tero and Aittala, Miika and Aila, Timo and Laine, Samuli},
  journal={Adv. Neural Inform. Process. Syst.},
  volume={35},
  pages={26565--26577},
  year={2022}
}

@article{song2021maximum,
  title={Maximum likelihood training of score-based diffusion models},
  author={Song, Yang and Durkan, Conor and Murray, Iain and Ermon, Stefano},
  journal={Adv. Neural Inform. Process. Syst.},
  volume={34},
  pages={1415--1428},
  year={2021}
}

@inproceedings{song2020score,
  title={Score-based generative modeling through stochastic differential equations},
  author={Song, Yang and Sohl-Dickstein, Jascha and Kingma, Diederik P and Kumar, Abhishek and Ermon, Stefano and Poole, Ben},
  booktitle={Int. Conf. Learn. Represent.},
  year={2021}
}

@article{lu2022dpm,
  title={{DPM-Solver}: A fast {ODE} solver for diffusion probabilistic model sampling in around 10 steps},
  author={Lu, Cheng and Zhou, Yuhao and Bao, Fan and Chen, Jianfei and Li, Chongxuan and Zhu, Jun},
  journal={Adv. Neural Inform. Process. Syst.},
  volume={35},
  pages={5775--5787},
  year={2022}
}

@article{zhao2023unipc,
  title={{UniPC}: A unified predictor-corrector framework for fast sampling of diffusion models},
  author={Zhao, Wenliang and Bai, Lujia and Rao, Yongming and Zhou, Jie and Lu, Jiwen},
  journal={Adv. Neural Inform. Process. Syst.},
  volume={36},
  pages={49842--49869},
  year={2023}
}

@inproceedings{chung2024cfgpp,
  title={{CFG++}: Manifold-constrained classifier free guidance for diffusion models},
  author={Chung, Hyungjin and Kim, Jeongsol and Park, Geon Yeong and Nam, Hyelin and Ye, Jong Chul},
  booktitle={Int. Conf. Learn. Represent.},
  year={2025}
}

@inproceedings{cao2023masactrl,
  title={{MasaCtrl}: Tuning-free mutual self-attention control for consistent image synthesis and editing},
  author={Cao, Mingdeng and Wang, Xintao and Qi, Zhongang and Shan, Ying and Qie, Xiaohu and Zheng, Yinqiang},
  booktitle={IEEE Conf. Comput. Vis. Pattern Recog.},
  pages={22560--22570},
  year={2023}
}

@inproceedings{parmar2023zero,
  title={Zero-shot image-to-image translation},
  author={Parmar, Gaurav and Kumar Singh, Krishna and Zhang, Richard and Li, Yijun and Lu, Jingwan and Zhu, Jun-Yan},
  booktitle={ACM SIGGRAPH 2023 Conf. Proc.},
  pages={1--11},
  year={2023}
}

@InProceedings{Rombach_2022_CVPR,
  author    = {Rombach, Robin and Blattmann, Andreas and Lorenz, Dominik and Esser, Patrick and Ommer, Bj\"orn},
  title     = {High-Resolution Image Synthesis With Latent Diffusion Models},
  booktitle = {IEEE Conf. Comput. Vis. Pattern Recog.},
  year      = {2022},
  pages     = {10684-10695}
}

@inproceedings{sdxl,
 author = {Podell, Dustin and English, Zion and Lacey, Kyle and Blattmann, Andreas and Dockhorn, Tim and M\"{u}ller, Jonas and Penna, Joe and Rombach, Robin},
 booktitle = {Int. Conf. Learn. Represent.},
 pages = {1862--1874},
 title = {SDXL: Improving Latent Diffusion Models for High-Resolution Image Synthesis},
 year = {2024}
}

@InProceedings{LPIPS,
    author = {Zhang, Richard and Isola, Phillip and Efros, Alexei A. and Shechtman, Eli and Wang, Oliver},
    title = {The Unreasonable Effectiveness of Deep Features as a Perceptual Metric},
    booktitle = {IEEE Conf. Comput. Vis. Pattern Recog.},
    year = {2018}
}

@ARTICLE{SSIM,
    author={Zhou Wang and Bovik, A.C. and Sheikh, H.R. and Simoncelli, E.P.},
    journal={IEEE Trans. Image Process.}, 
    title={Image quality assessment: from error visibility to structural similarity}, 
    year={2004},
    volume={13},
    number={4},
    pages={600-612},
}

@inproceedings{CLIP,
  title={Learning transferable visual models from natural language supervision},
  author={Radford, Alec and Kim, Jong Wook and Hallacy, Chris and Ramesh, Aditya and Goh, Gabriel and Agarwal, Sandhini and Sastry, Girish and Askell, Amanda and Mishkin, Pamela and Clark, Jack and others},
  booktitle={Int. Conf. Mach. Learn.},
  pages={8748--8763},
  year={2021},
}

@inproceedings{MSCOCO,
  title={Microsoft {COCO}: Common objects in context},
  author={Lin, Tsung-Yi and Maire, Michael and Belongie, Serge and Hays, James and Perona, Pietro and Ramanan, Deva and Doll{\'a}r, Piotr and Zitnick, C Lawrence},
  booktitle={Eur. Conf. Comput. Vis.},
  pages={740--755},
  year={2014},
}

@inproceedings{nocaps,
  title={{NOCAPS}: novel object captioning at scale},
  author={Agrawal, Harsh and Desai, Karan and Wang, Yufei and Chen, Xinlei and Jain, Rishabh and Johnson, Mark and Batra, Dhruv and Parikh, Devi and Lee, Stefan and Anderson, Peter},
  booktitle={Int. Conf. Comput. Vis.},
  pages={8948--8957},
  year={2019}
}

@article{anderson1965iterative,
  title={Iterative procedures for nonlinear integral equations},
  author={Anderson, Donald G},
  journal={Journal of the ACM (JACM)},
  volume={12},
  number={4},
  pages={547--560},
  year={1965},
  publisher={ACM New York, NY, USA}
}

@article{lin2024schedule,
  title={Schedule your edit: A simple yet effective diffusion noise schedule for image editing},
  author={Lin, Haonan and Chen, Yan and Wang, Jiahao and An, Wenbin and Wang, Mengmeng and Tian, Feng and Liu, Yong and Dai, Guang and Wang, Jingdong and Wang, Qianying},
  journal={Adv. Neural Inform. Process. Syst.},
  volume={37},
  pages={115712--115756},
  year={2024}
}

@ARTICLE{TMM10814072,
  author={Yu, Hua and Hou, Yaqing and Pei, Wenbin and Ong, Yew-Soon and Zhang, Qiang},
  journal={IEEE Transactions on Multimedia}, 
  title={{DivDiff}: A Conditional Diffusion Model for Diverse Human Motion Prediction}, 
  year={2025},
  volume={27},
  pages={1848-1859},
}

@inproceedings{hong2024exact,
  title={On exact inversion of {DPM-Solver}s},
  author={Hong, Seongmin and Lee, Kyeonghyun and Jeon, Suh Yoon and Bae, Hyewon and Chun, Se Young},
  booktitle={IEEE Conf. Comput. Vis. Pattern Recog.},
  pages={7069--7078},
  year={2024}
}

@InProceedings{wallace2023edict,
    author    = {Wallace, Bram and Gokul, Akash and Naik, Nikhil},
    title     = {{EDICT}: Exact Diffusion Inversion via Coupled Transformations},
    booktitle = {IEEE Conf. Comput. Vis. Pattern Recog.},
    year      = {2023},
    pages     = {22532-22541}
}

@inproceedings{zhang2024exact,
  title={Exact diffusion inversion via bidirectional integration approximation},
  author={Zhang, Guoqiang and Lewis, Jonathan P and Kleijn, W Bastiaan},
  booktitle={Eur. Conf. Comput. Vis.},
  pages={19--36},
  year={2024},
}

@article{wang2024belm,
  title={{BELM}: Bidirectional explicit linear multi-step sampler for exact inversion in diffusion models},
  author={Wang, Fangyikang and Yin, Hubery and Dong, Yue-Jiang and Zhu, Huminhao and Zhao, Hanbin and Qian, Hui and Li, Chen and others},
  journal={Adv. Neural Inform. Process. Syst.},
  volume={37},
  pages={46118--46159},
  year={2024}
}

@InProceedings{Zhou_2025_Golden,
    author    = {Zhou, Zikai and Shao, Shitong and Bai, Lichen and Zhang, Shufei and Xu, Zhiqiang and Han, Bo and Xie, Zeke},
    title     = {Golden Noise for Diffusion Models: A Learning Framework},
    booktitle = {Int. Conf. Comput. Vis.},
    year      = {2025},
    pages     = {17688-17697}
}

@inproceedings{bai2024zigzag,
  title={Zigzag Diffusion Sampling: Diffusion Models Can Self-Improve via Self-Reflection},
  author={Bai, Lichen and Shao, Shitong and Zhou, Zikai and Qi, Zipeng and Xu, Zhiqiang and Xiong, Haoyi and Xie, Zeke},
  booktitle={Int. Conf. Learn. Represent.},
  year={2025}
}
\bibliographystyle{IEEEtran}

\vfill

\end{document}